\documentclass{article}

\usepackage{arxiv}

\usepackage[utf8]{inputenc} 
\usepackage[T1]{fontenc}    
\usepackage{hyperref}       
\usepackage{url}            
\usepackage{booktabs}       
\usepackage{amsfonts}       
\usepackage{nicefrac}       
\usepackage{microtype}      
\usepackage{lipsum}
\usepackage{graphicx}

\usepackage{algorithm2e}

\usepackage{booktabs}   
\usepackage{makecell}   

\usepackage{tabularx}  
\usepackage{booktabs}  
\usepackage{array}     

\usepackage{amsthm}

\theoremstyle{definition}

\usepackage{epsfig}
\usepackage{subcaption}
\usepackage{calc}
\usepackage{amssymb}
\usepackage{amstext}
\usepackage{amsmath}
\usepackage{amsthm}
\usepackage{multicol}
\usepackage{pslatex}
\usepackage{apalike}
\usepackage{algorithm2e}
\usepackage[bottom]{footmisc}

\usepackage{enumitem}   
\usepackage{hyperref}   

\graphicspath{ {./images/} }

\title{HADA: Human-AI Agent Decision Alignment Architecture}

\author{
  Tapio Pitkäranta \\
  Department of Computer Science and Engineering\\
  Aalto University, 
  Finland \\
  \texttt{tapio.pitkaranta@iki.fi} \\
  \And
  Leena Pitkäranta \\
  Department of Industrial Engineering and Management\\
  Aalto University, 
  Finland \\
  \texttt{leena.pitkaranta@iki.fi} \\
}

\begin{document}
\maketitle

\begin{abstract}
    \textbf{Problem \& Motivation.} 
    The \emph{generative AI boom} is spawning rapid deployment of diverse \emph{LLM software agents}.  
    New standards such as the \emph{Model Context Protocol} (MCP) and \emph{Agent-to-Agent (A2A)} protocols let agents share data and tasks, yet organizations still lack a rigorous way to keep those agents — and legacy algorithms — aligned with organizational targets and values.  
      
    \textbf{Objectives of the Solution.}
    We aim to deliver a software reference architecture that
    (i) provides every stakeholder natural-language interaction across planning horizons with software agents and AI algorithmic logic,
    (ii) provides a multi-dimensional way for aligning stakeholder targets and values with algorithms and agents,
    (iii) provides an example for jointly modelling AI algorithms, software agents, and LLMs,
    (iv) provides a way for stakeholder interaction and alignment across time scales,
    (v) scales to thousands of algorithms and agents while remaining auditable,
    (vi) remains \emph{framework-agnostic}, allowing the use of any underlying LLM, agent library, or orchestration stack without requiring redesign.

    \textbf{Design \& Development.} 
    Guided by the \emph{Design-Science Research Methodology (DSRM)}, we engineered \emph{HADA} (Human-Algorithm Decision Alignment)—a protocol-agnostic, multi-agent architecture that layers role-specific interaction agents over both Large-Language Models and legacy decision algorithms.  
    Our reference implementation containerises a production credit-scoring model, \texttt{getLoanDecision}, and exposes it through stakeholder agents (business manager, data scientist, auditor, ethics lead and customer), enabling each role to steer, audit and contest every decision via natural-language dialogue.  
    The resulting constructs, design principles and justificatory knowledge are synthesised into a mid-range design theory that generalises beyond the banking pilot.  
    
    \textbf{Demonstration.} 
    HADA is instantiated on a cloud-native stack—\emph{Docker, Kubernetes} and \emph{Python}—and embedded in a retail-bank sandbox.  
    Five scripted scenarios show how business targets, algorithmic parameters, decision explanations and ethics triggers propagate end-to-end through the HADA architecture.
    
    \textbf{Evaluation.} 
    Walkthrough observation and log inspection were used to gauge HADA against six predefined objectives.  
    A stakeholder–objective coverage matrix showed 100 \% fulfilment: every role could invoke conversational control, trace KPIs and values, detect and correct bias (ZIP-code case), and reproduce decision lineage—without dependence on a particular agent hierarchy or LLM provider.  
    
    \textbf{Contributions.} 
    The research delivers (i) an open-source HADA reference architecture, (ii) an evaluated mid-range design theory for human–AI alignment in multi-agent settings, and (iii) empirical evidence that framework-agnostic, protocol-compliant stakeholder agents can simultaneously enhance accuracy, transparency and ethical compliance in real-world decision pipelines.

\end{abstract}




\section{Introduction}
\label{sec:introduction}

The recent surge of transformer-based Large Language Models (LLMs)\mbox{—}from BERT~\cite{devlin2018bert} through GPT-3~\cite{brown2020language} to GPT-4—has turned artificial intelligence from an expert playground into a ubiquitous, conversational medium embedded in e-mail clients, code editors and service chatbots.  
Powered by the self-attention mechanism introduced in the seminal Transformer architecture~\cite{vaswani2017attention}, modern LLMs let non-experts question, instruct and refine algorithms in plain language, shifting human–machine interaction from rigid menu clicks to rich, bidirectional dialogue.  
This democratisation is fuelling an unprecedented wave of LLM software \emph{agents}, as teams wrap models with tool-calling, planning and memory modules to automate tasks across every industry sector.  Notably, some LLMs have reportedly \emph{passed the Turing test}, underscoring their ability to mimic human-level conversation~\cite{jones2025largelanguagemodelspass}.  
As organisations embed these self-improving, multi-agent systems in mission-critical workflows, they hand over growing autonomy—and thus responsibility—to algorithmic decision makers.  When that autonomy drifts from stakeholder values, the results can resemble the “weapons of math destruction” critics have warned about~\cite{o2017weapons}.  Solving this \emph{alignment problem} requires a design-science perspective that marries technical advances with managerial and ethical safeguards~\cite{christian2020alignment,wilson2000consilience}.  


We propose \emph{HADA}—\textbf{H}uman-\textbf{A}I Agent \textbf{D}ecision \textbf{A}lignment. HADA wraps a single production AI Tools in stakeholder-specific agents, enabling business managers, data scientists, auditors, ethics leads and customers to steer, audit and contest every credit decision through dialogue.  While the pilot centres on this banking use-case, HADA’s metamodel is protocol-agnostic: it natively supports the Model Context Protocol (MCP) and emerging Agent-to-Agent (A2A) standards, and remains compatible with any LLM, agent framework or orchestration stack. HADA builds on earlier Retrieval-Augmented Generation prototype that exposed natural-language controls to business users~\cite{pitkaranta2024bridging}.

\subsection{Research Methodology}
This study follows the Design Science Research methodology (DSR) to bridge theory and practice through iterative artifact creation and evaluation.  
Guided by the canonical DSR process model~\cite{peffers2020designscienceresearchprocess}, we executed problem identification, objective definition, artefact design, demonstration and evaluation cycles.  
Consistent with Gregor and Jones’s anatomy of mid-range design theory~\cite{gregor2007anatomy} and the foundational DSR guidelines by Hevner et al.~\cite{hevner2004design}, we formalised HADA’s design rationale, constructs and principles as a \emph{mid-range theory}—general enough to transfer across domains, yet concrete enough to guide engineering practice.
Large language models were used to assist with spell checking\footnote{During the preparation of this manuscript, the authors made limited use of several OpenAI generative-AI services—namely \textit{GPT-4o}, \textit{GPT-o3}, and \textit{GPT-o4-mini}—exclusively for language polishing (grammar, stylistic refinement, typo correction) and reference-list formatting. All prompts, intermediate outputs, and final wording were critically reviewed and verified by the authors, who accept full responsibility for the content.}.

\subsection{Problem Identification}
Despite rapid uptake of LLM agents and legacy decision algorithms, three persistent gaps hinder responsible deployment:  

\begin{enumerate}[label=\textbf{G\arabic*}, leftmargin=3em]
    \item \textbf{Alignment Gap} — Organisations lack systematic mechanisms to keep both algorithms and agents aligned with measurable KPIs, codified values and evolving regulation~\cite{christian2020alignment,amodei2016concrete}.  
    \item \textbf{Containment Gap} — Beyond objective alignment, enterprises still lack reliable kill-switches, circuit breakers and anomaly detection that can restrain autonomous systems when they deviate or self-modify in production~\cite{Suleyman2023coming,yampolskiy2014containment,brundage2020toward,hadfieldmenell2016offswitch}.
    \item \textbf{Integration Gap} — Although patterns such as Retrieval-Augmented Generation exist, practitioners have few vendor-agnostic blueprints for embedding existing decision algorithms into multi-agent, LLM-powered workflows while maintaining end-to-end auditability~\cite{pitkaranta2024bridging,lewis2020retrieval}.  
\end{enumerate}

\subsection{Objectives of the Solution}
\label{subsec:objectives}
HADA addresses these gaps through six design objectives:

\begin{enumerate}[label=\textbf{O\arabic*}, leftmargin=3em]
  \item \textbf{Natural-language interaction across planning horizons} — Equip every stakeholder—from C-suite to on-call engineers—with a conversational interface that lets them restate or define \emph{quarterly} Objectives and Key Results (OKRs) and have those changes automatically translated into concrete, \emph{daily} algorithmic parameters and guard-rails.
  \item \textbf{Multi-dimensional alignment (targets \& values)} — Provide two complementary tracks:  
        \emph{(i) OKR Alignment}: bind each AI algorithm’s reward function or decision logic to measurable Objectives \& Key Results;  
        \emph{(ii) Values Alignment}: encode organisational ethical principles as machine-readable policies and maintain a verifiable audit chain from high-level values to low-level actions.
  \item \textbf{Reference architecture for hierarchical, hybrid AI systems} — Provide clear definitions of LLM agents and AI algorithms, and address the AI containment gap. Specify a modular stack that separates the strategy layer (responsible for goal and policy management), the coordination layer (responsible for multi-agent task brokerage), and the execution layer (comprising LLM agents and classical algorithms), so that revising a quarterly objective triggers an orchestrated cascade of updates while preserving value constraints.
  \item \textbf{Stakeholder alignment across time scales} — Provide a repeatable method for eliciting and reconciling stakeholder goals at annual, quarterly and operational cadences, then encoding them as machine-readable policies that guide daily agent behaviour.
  \item \textbf{Scalability, auditability and near-real-time propagation} — Support thousands of agents and algorithms while propagating objective or policy changes from strategy to production within hours, without sacrificing audit trails, explainability or rollback safety.
  \item \textbf{Framework-agnostic, policy-driven design} — Remain independent of any single LLM, agent framework or orchestration tool; represent objectives and policies in open standards (e.g.\ MCP, A2A) so that strategic edits flow unimpeded to heterogeneous runtime environments.
\end{enumerate}

\section{Related Work: Emerging LLM Software Agents}

\textit{LLM agents}—sometimes called \emph{LLM-powered agents}—combine a large language model with an explicit control loop.  Formally, such an agent maintains (i) a prompt or system specification that encodes its initial policy, (ii) a mutable conversation trace that acts as state, and (iii) the ability to \texttt{act} on an external environment via tools or APIs \cite{cemri2025multiagentllmsystemsfail,naveed2024comprehensiveoverviewlargelanguage}.  Leveraging the underlying LLM for perception and inference, the agent can perform natural-language understanding and generation, task decomposition, planning, decision making, and tool invocation \cite{xi2023rise,wang2023survey,yang2023foundation}.  Concretely, a user request may be broken into sub-tasks, relevant information fetched from the web, external services called, and the aggregated result summarised for the user \cite{yao2022react,shinn2023reflexion}.

LLM agents have progressed rapidly, yet several hurdles remain.  Hallucinations—plausible but incorrect answers—undermine reliability \cite{ji2023survey}, prompting mitigation via grounding and retrieval-augmented generation (RAG) \cite{lewis2020retrieval}.  Further obstacles include safety and ethical concerns, computational expense, and biases inherited from training data \cite{bender2021dangers,weidinger2021ethical}.  The opaque “black-box’’ reasoning of large models also hampers debugging and trust in high-stakes domains \cite{doshi2017towards}. Current research therefore targets higher factuality, better explainability and scalable multi-agent coordination.  Approaches such as web-grounded training \cite{nakano2021webgpt}, dynamic context selection \cite{ram2023context}, and richer evaluation benchmarks all seek to deliver agents that are both more accurate and more transparent—while emphasising responsible, ethics-aware deployment.

\subsection{Single LLM Agent}

\begin{figure*}[t]
    \centering
    \includegraphics[width=0.4\linewidth]{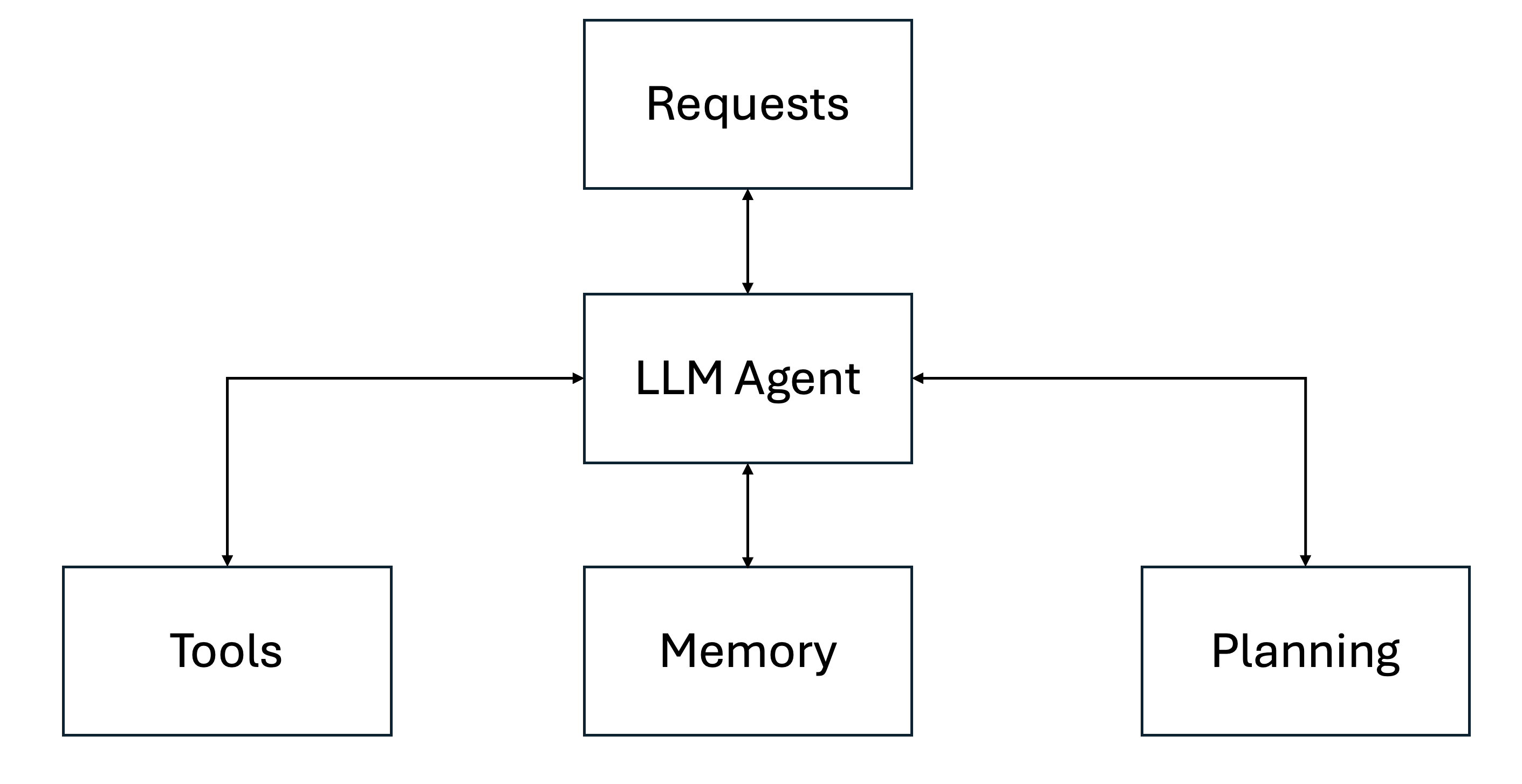}
    \caption{Single LLM agent core components}
    \label{fig:llm_agents}
\end{figure*}

The Figure~\ref{fig:llm_agents} illustrates a high-level architecture of an LLM Agent, which consists of the following components: \textbf{Requests}, which serve as input from a human user or another agent initiating the agent's operations; the \textbf{LLM Agent}, the central module that utilizes a Large Language Model (LLM) built upon a multimodal foundational model, acting as the coordinator and managing interactions with all other components; \textbf{Tools}, which are external modules or systems the agent can employ to gather data or perform specific tasks; \textbf{Memory}, a mechanism for storing and retrieving past interactions or knowledge, whether short-term or long-term; and \textbf{Planning}, a module responsible for devising a structured approach to solve complex tasks by breaking them down into subtasks. While some studies provide separate definitions for Actions and Tools \cite{yang2025surveyaiagentprotocols}, we group them under the unified \textbf{Tools} concept in this context.

\subsubsection{Memory}

The memory module stores the agent’s internal logs, including prior thoughts, actions, observations, and interactions with users. Memory is generally categorized into \textbf{short-term memory}, which captures immediate context and is limited by the LLM’s context window through in-context learning; \textbf{long-term memory}, which stores persistent knowledge and past experiences using external vector databases for scalable retrieval; and \textbf{hybrid memory}, which combines short-term and long-term memory to enhance reasoning and accumulate experiences over time.

\subsubsection{Tools and Actions}

Tools enable LLM agents to take actions and interact with external systems such as APIs, databases, knowledge bases, and external models. They support workflows that help the agent gather information or complete subtasks, like using a code interpreter to generate charts in response to user queries \cite{naveed2024comprehensiveoverviewlargelanguage}.

\subsubsection{Planning} 

The \emph{planning module} decomposes a user’s high-level goal into a sequence of smaller, tractable subtasks that the agent can address in turn.  This explicit task-structure not only guides downstream execution but also gives the system a scaffold for more reliable reasoning. \textbf{Planning without Feedback.} When the plan is created in a single pass, task decomposition relies on internal chains of reasoning generated by an LLM.  Two widely cited paradigms are \emph{Chain-of-Thought} and \emph{Tree-of-Thoughts}, representing single-path and multi-path reasoning, respectively \cite{Wang_2024}.

\textbf{Planning with Feedback.} One-shot plans struggle with long-horizon or error-prone tasks because they lack a mechanism to correct earlier decisions.  Feedback-driven approaches overcome this limitation by iteratively revising the plan in light of previous actions and observations.  Among the most influential are \emph{ReAct} \cite{yao2023reactsynergizingreasoningacting} and \emph{Reflexion} \cite{shinn2023reflexionlanguageagentsverbal}. \emph{ReAct} interleaves three repeated steps—\textbf{Thought}, \textbf{Action}, and \textbf{Observation}—so that each new reasoning step conditions on feedback from the environment, optional human critiques, or model-based self-evaluation \cite{yao2023reactsynergizingreasoningacting}.  In contrast, \emph{Reflexion} augments an agent with explicit \textbf{Evaluator} and \textbf{Self-reflection} LMs plus short- and long-term memory buffers.  After every episode, linguistic feedback is stored in an episodic memory that shapes future decisions, enabling continual improvement across coding, reasoning, and decision-making tasks \cite{shinn2023reflexionlanguageagentsverbal}.

\subsection{Emerging LLM Agent Protocols} 

A chronological review of recent work shows that the first protocols used by LLM agents surfaced in 2024 \cite{yang2025surveyaiagentprotocols}.  Initial specifications such as Agent Network Protocol (\emph{ANP}) and Model Context Protocol (\emph{MCP}) tackled the absence of standard interfaces that had limited large-language-model (LLM) agents’ capacity to scale, interoperate, and connect with external tools and data sources.  This groundwork was refined by lighter-weight formats—most notably \emph{ACP}, Google’s \emph{A2A} (Agent-to-Agent) protocol, and \texttt{agents.json}—together with infrastructure projects such as \emph{Agora}, all aimed at richer, more structured messaging between agents.  The overall trajectory can be grouped into three phases: early ad-hoc standards, consolidation around concise schemas, and the forthcoming generation of layered specifications.  Looking ahead to 2025 and beyond, initiatives such as \emph{OpenManus} and \emph{OWL} intend to formalise group-centric, multi-layer protocols, signalling a transition toward fully interoperable, real-world agent ecosystems \cite{yang2025surveyaiagentprotocols}.

\subsubsection{Model Context Protocol (MCP): LLM–Tools Integration}

The \emph{Model Context Protocol} (MCP) is an open-source standard that exposes real-time data from content-management, enterprise, and developer systems to LLM-based assistants\footnote{\url{https://modelcontextprotocol.io/introduction}}.  In Figure~\ref{fig:llm_agents} it functions as the middleware layer between the LLM Agent and external Tools. By replacing ad-hoc, one-off connectors with a single interface, MCP eliminates most of the bespoke plumbing normally required for each new data source.  The result is simpler integration, live context for the model, and more accurate, up-to-date outputs—all while letting developers concentrate on higher-level application logic rather than low-level data wiring.

At its core, MCP follows a client-server architecture where a host application can connect to multiple servers. The architecture includes \textbf{MCP Hosts}, which are programs, IDEs, AI tools, or LLM Agents that want to access data through MCP; \textbf{MCP Clients}, which are protocol clients that maintain 1:1 connections with servers; \textbf{MCP Servers}, which are lightweight programs that each expose specific capabilities through the standardized Model Context Protocol; \textbf{Local Data Sources}, which refer to your computer’s files, databases, and services that MCP servers can securely access; and \textbf{Remote Services}, which are external systems available over the internet (e.g., through APIs) that MCP servers can connect to.

\subsubsection{Google Agent2Agent Protocol (A2A)}

Agent2Agent (A2A) is an open protocol by Google that aims to complement the Model Context Protocol (MCP), which provides helpful tools and context to agents \footnote{https://developers.googleblog.com/en/a2a-a-new-era-of-agent-interoperability/}. The A2A protocol will allow AI agents to communicate with each other, securely exchange information, and coordinate actions on top of various enterprise platforms or applications. In Figure~\ref{fig:llm_agents} it functions as the middleware layer between two LLM Agents interacting with each other.

\textbf{Conceptual Overview:} The Agent2Agent (A2A) protocol facilitates communication between independent AI agents through several key elements\footnote{https://github.com/google/A2A}. These include the \textbf{Agent Card}, which is public JSON metadata detailing an agent's capabilities, endpoint, and authentication for discovery; the \textbf{A2A Server}, which is an agent exposing an HTTP endpoint implementing A2A protocol methods to receive requests and manage tasks; the \textbf{A2A Client}, which is an application or agent consuming A2A services by sending requests (e.g., tasks/send) to a server URL; the \textbf{Task}, which is the core work unit initiated by a client message (tasks/send or tasks/sendSubscribe), with a unique ID and lifecycle states (submitted, working, input-required, completed, failed, canceled); and the \textbf{Message}, which represents a communication turn between client ("user") and agent ("agent"), containing Parts. A \textbf{Part} is the basic content unit in a Message or Artifact, such as TextPart, FilePart (inline or URI), or DataPart (structured JSON); an \textbf{Artifact} refers to agent-generated outputs during a task (e.g., files, structured data), also composed of Parts. For long tasks, \textbf{Streaming} is supported through tasks/sendSubscribe, enabling servers to send real-time progress via Server-Sent Events (TaskStatusUpdateEvent or TaskArtifactUpdateEvent). Additionally, \textbf{Push Notifications} allow servers with pushNotifications enabled to proactively send task updates to a client-provided webhook URL, which is configured via tasks/pushNotification/set.

\subsection{Multi-Agent Conversation Patterns in LLM Era}

Multi-agent frameworks such as \emph{AutoGen} show that once more than two LLM agents collaborate, simple one-to-one exchanges become a bottleneck; the communication graph must evolve from isolated dialogues into richer, group-level patterns \cite{wu2023autogenenablingnextgenllm}.  In a \emph{joint-chat} scheme, all agents share a common channel, allowing rapid knowledge propagation but also demanding mechanisms for turn-taking and conflict resolution.  Alternatively, a \emph{hierarchical-chat} topology appoints one or more “manager’’ agents that coordinate a set of specialised workers, reducing the $O(n^{2})$ message overhead of full pairwise gossip to roughly $O(n)$ while preserving global context.  The general lesson, echoed by the schematic in the accompanying image, is that scalable systems require flexible conversation patterns—peer-to-peer for small teams, hub-and-spoke or tree-structured for larger cohorts—to balance reasoning depth with communication cost.

\subsection{Prominent LLM Agent Frameworks and Libraries}

A fast-maturing ecosystem of open-source frameworks now captures the canonical design patterns for
agentic large-language-model (LLM) systems, letting researchers focus on reasoning strategies rather
than boiler-plate glue code.  \textbf{LangChain} supplies composable primitives—tools, memories and
chains—that underpin many production chatbots, document-question-answering (QA) pipelines and research
assistants.  Building on the same abstractions, \textbf{LangGraph} introduces a graph-based execution
engine with explicit state, branching and parallelism, which is invaluable for long-horizon planning
and multi-agent coordination.

Microsoft’s \textbf{AutoGen} formalises ``chat-loop’’ interactions in which several LLMs, optionally
augmented by external tools or human feedback, iteratively critique and refine one another to solve
non-trivial tasks.  Where private or proprietary data are involved, \textbf{LlamaIndex} offers a
complete data layer—index construction, retrieval interfaces and knowledge-graph support—for
retrieval-augmented generation (RAG) over local corpora.  \textbf{CrewAI} targets role-oriented
workflows: agents endowed with complementary expertise share a common memory to tackle coordinated
problem solving or simulation environments. For production use, Microsoft’s \textbf{Semantic Kernel} provides a language-agnostic plug-in system,
a symbolic planner and robust security controls that simplify the integration of generative AI into
enterprise software stacks.  Google’s \textbf{Agent Development Kit} (ADK) implements the
Agent-to-Agent (A2A) protocol and offers orchestration, safety guards and evaluation utilities for
multi-agent research at scale.  Also from Google, \textbf{Genkit} delivers a code-first, TypeScript
workflow with flow orchestration, prompt/version control (\texttt{.prompt} files), observability and
one-command Cloud Run deployment.  Finally, \textbf{Marvin} provides a pythonic decorator-based API
that turns ordinary functions into threaded, asynchronous ``AI functions,’’ making rapid prototyping
inside data-science or machine-learning pipelines straightforward. Additional libraries—including Haystack, 
Embedchain, SuperAGI, Dify and OpenAgents—continue to expand the design space and keep the agent-framework 
landscape highly dynamic.

\section{Design \& Development}
\label{sec:design-development}

This section instantiates the Design Science Research Methodology (DSRM) \cite{peffers2020designscienceresearchprocess} step~3 (``\emph{Design \& Development}'') by describing the three concrete artifacts that operationalise the HADA concept.  Together they realise the solution objectives defined in Section~\ref{sec:introduction}.

\paragraph{Context.}
The pilot exposes the \emph{alignment problem} that unfolds across the
entire life-cycle of an AI-driven credit-approval workflow in a retail bank.  
The cycle is triggered by a strategic shift in business targets, proceeds
through model redesign and validation, and culminates in production
deployment—each phase demanding tight coordination among heterogeneous
roles.  
While the organisation strives to shorten ``time-to-yes'' for unsecured
consumer credit, it must simultaneously comply with Basel III, the EU AI Act
and local consumer-protection statutes.  
Operational responsibility for every credit decision is delegated to
\texttt{getLoanDecision}, an inherently explainable decision-tree model
currently running in production.  

To bridge the gaps that arise between technical optimisation, ethical
commitments and regulatory oversight along this end-to-end trajectory, we
propose \emph{HADA} (Human-Aligned Decision Alignment).  
HADA not only orchestrates the activities of multiple stakeholders—adding
new governance roles (e.g.\ a \emph{Value and Ethics Manager}) and expanding
the remit of existing ones (e.g.\ the \emph{Audit Manager})—but also
delivers a complementary suite of AI-architecture tools, such as a
version-controlled feature catalogue, a transparent model registry, an
ethics-trigger engine and an immutable decision ledger.  
These technical components furnish each role with the information and
control primitives required to discharge its responsibilities, thereby
addressing alignment challenges that traditional score-card systems leave
unresolved.

\subsubsection*{Structure of the artifacts}
The \emph{Design \& Development} section is organised around six inter-locking artifacts: \textbf{Artifact~\ref{ssec:artifact-strategy} — Strategy-Process Blueprint}. Formalises the yearly $\rightarrow$ daily planning loop (OKRs, KPI cascades, operational reviews) that frames every subsequent decision governed by HADA; \textbf{Artifact~\ref{ssec:artifact-stakeholder-user-story} — Stakeholder \& User-Story Specification}. Captures the problem context through a stakeholder map, sequenced user stories, and an extended RACI matrix using Artifact~\ref{ssec:artifact-getLoanDecision} as concrete AI Tool / AI Tool; \textbf{Artifact~\ref{ssec:artifact-getLoanDecision} — Prototype AI Tool \texttt{getLoanDecision}}. An operational decision-tree model (dataset, feature engineering, API) that instantiates the credit-approval use-case; \textbf{Artifact~\ref{ssec:artifact-RACI} — AI-Responsibility Tool (RACI Matrix)}. Distils governance duties across eight life-cycle activities, ensuring accountable, auditable hand-offs between roles regarding Artifact~\ref{ssec:artifact-getLoanDecision}; \textbf{Artifact~\ref{ssec:artifact-hada-architecture} — Generic HADA Architecture}. Provides the layered, protocol-agnostic reference design (agent layer, tools layer, DevOps view) on which all domain instances rest; \textbf{Artifact~\ref{ssec:artifact-agent-modelling} — Agent Modelling}. Maps human roles to containerised interaction agents and describes the HADA Controller pattern for policy-enforced orchestration regarding.

\subsection{Artifact -- Strategy-Process Blueprint}
\label{ssec:artifact-strategy}

\begin{figure}[t]
    \centering
    \includegraphics[width=0.9\textwidth]{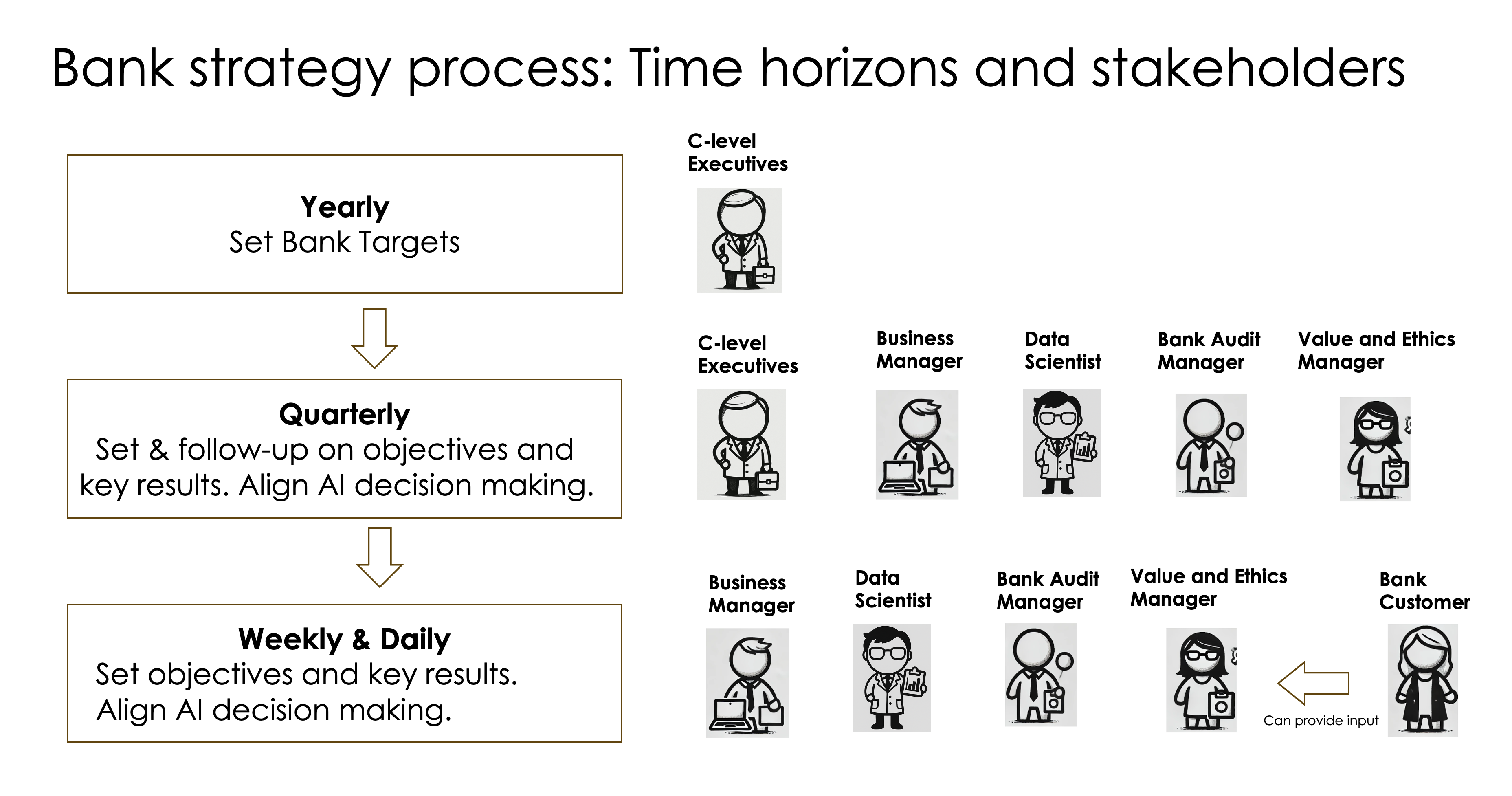}
    \caption{Bank strategy process: time horizons and stakeholder coverage}
    \label{fig:bank-strategy-process}
\end{figure}

\paragraph{Purpose.}
Artifact~\ref{ssec:artifact-strategy} formalises the yearly-to-daily planning loop that steers our bank from strategical decisions to daily activities as illustrated in Figure~\ref{fig:bank-strategy-process}. Our example bank follows a typical strategy formulation and implementation process in which the yearly overall targets are set in the beginning of the year and the progress and continuous alignment of the more detailed activities is followed up typically at least on monthly or quarterly basis. One popular example of such a framework is 'Objectives and Key Results' (OKR) framework \cite{niven_objectives_2016}. 


The OKR framework formalizes the key results follow-up process and describes the mechanisms required to create alignment in the organization. While the C-level executives are responsible for the overall strategy formulation and set the high-level objectives for the organization, the experts and managers on the lower organizational levels play a crucial role in implementing the strategy. Their responsibility is to ensure that the day-to-day decisions are aligned with the values and overall objectives of the organization and that the key results will be achieved as planned. In our bank example the ‘connecting OKRs’ process described by \cite{niven_objectives_2016} needs to be expanded to cover not only the decisions and actions taken by the human employees but also the algorithms.

\paragraph{Yearly — Set bank targets.}
At the start of each fiscal year the Chief Credit Officer (CCO) and peer C-level executives establish portfolio-level Objectives \& Key Results (OKRs) together with an updated risk-appetite statement. These artifacts are stored in the \emph{Business-Target Catalogue} and become immutable reference points for all downstream KPI specifications.

\paragraph{Quarterly — Translate OKRs into actionable KPIs.}
Every quarter the CCO reconvenes with Business Managers, Data Scientists, Audit Managers and Ethics Managers to review recent performance and—if necessary—re-weight model KPIs (e.g.\ shifting focus from “new-customer acquisition’’ to “expected-loss minimisation’’). 

\paragraph{Weekly \& Daily — Operative alignment loops.} On a rolling basis the following stakeholders contribute: Business Manager – reviews live KPI dashboards and reprioritises tickets; Data Scientist – publishes hot-fix model versions or feature toggles via the Model Catalogue; Audit Manager – samples recent decisions from the immutable ledger and verifies lineage; Ethical-Compliance Manager – inspects the sensitive-attribute watch-list and issues ethics triggers when required.

\paragraph{Contribution.} The strategy-process blueprint provides a single, auditable spine that a) anchors KPI and value alignment across all time horizons, b) delineates clear hand-offs between business, technical and oversight roles, and c) remains tooling-agnostic—no assumptions are made about the underlying LLM, agent library or orchestration layer.

\subsection{Artifact -- Stakeholder \& User-Story Specification}
\label{ssec:artifact-stakeholder-user-story}

Artifact~\ref{ssec:artifact-stakeholder-user-story} formalises the \emph{problem context} and the
\emph{functional requirements}.
It is released as a lightweight specification composed of  
(i) a stakeholder map,  
(ii) an ordered set of user stories, and  
(iii) a RACI matrix that clarifies the roles and responsibilities of each
stakeholder.
Together these elements provide a transparent overview of the actors
involved in the banking use-case and their interactions with the HADA
system.

Table~\ref{tab:stakeholders} enumerates the principal stakeholder roles,
articulates their primary goals and specifies the access rights granted by
HADA.  
The mapping was validated in three semi-structured workshops with
subject-matter experts from credit-risk, compliance and data-science
departments, reinforcing its credibility within the banking sector.

\subsubsection{Stakeholder map}
Table~\ref{tab:stakeholders} lists the five principal roles involved in the experimental banking scenario, their primary goals and the HADA components they are allowed to access.  


\begin{table}[h!]
    \centering
    \caption{Stakeholder map for the banking use-case \texttt{getLoanDecision}}
    \label{tab:stakeholders}
    \begin{tabular}{|p{2cm}|p{6cm}|p{6cm}|}
        \hline
        \textbf{Role} & \textbf{Primary goal in workflow} & \textbf{Key HADA privileges} \\ \hline
        \textbf{Customer} & Receive a fair, fast and transparent credit decision; raise complaints when needed & Submit loan-application data; obtain natural-language decision explanations; open ethics tickets \\[1ex] \hline
        \textbf{Chief Credit Officer (CCO)} & Define company-level Objectives \& Key Results (OKRs) for credit strategy; align risk appetite, growth targets and regulatory limits across the portfolio & Set and update top-level KPI/OKR targets; approve credit policy changes; view enterprise-wide risk dashboards; give final sign-off on new production decision AI Tool deployments \\[1ex] \hline
        \textbf{Business Manager} & Set and adjust portfolio KPIs (e.g.\ minimise expected credit losses) & Edit KPI targets; approve or reject AI Tool versions for production; view live KPI dashboards \\[1ex] \hline
        \textbf{Data Scientist} & Build, test and monitor the \texttt{getLoanDecision} model so it meets current KPI and risk targets & Retrain models; run offline validation; register new AI Tool versions in the Model Catalogue \\[1ex] \hline
        \textbf{Audit Manager} & Verify compliance and traceability of any individual credit decision & Read-only access to the full decision path, model version, feature values and audit ledger \\[1ex] \hline
        \textbf{Value \& Ethics Manager} & Safeguard ethical use of data and enforce organisational values across all AI Tools & Maintain Values \& Data catalogues; flag or deprecate sensitive attributes; approve/reject attribute changes; mandate model retraining when catalogue updates occur \\ \hline
    \end{tabular}
\end{table}

\subsubsection{User stories}

User stories used in this implementation described in the below itemized list. The user stories are sequential: first business KPI's are changed by one stakeholder and AI Tool optimized to achieve new targets. This phase induces ethically questionable logic into the AI Tool but it is not noted by the stakeholder. Then customer applies for loan and gets the loan with questionable logic. Then customer returns to file a complaint as he / she noted that questionnable data was used in the loan decision. Then HADA system contacts auditor and after that value and ethics manager to adjust the system.


\begin{itemize}
    \item As the \textbf{Chief Credit Officer (CCO)}, I want to update the quarterly OKRs from acquiring new customers to minimising credit losses during the annual strategy-planning process.

    \item As a \textbf{Bank Loan Department Business Manager}, I would like to change the business target for short-term loan decisions from acquiring new customers to minimising credit problems.

    \item As a \textbf{Data Scientist}, I want to create a new version of the \texttt{getLoanDecision} AI Tool so that it matches the new business targets.

    \item As a \textbf{Bank Customer}, I would like to apply for a short-term loan.

    \item As a \textbf{Bank Customer}, I would like to file a complaint with the ethical value-alignment, because I am being asked for ethically questionable information while applying for a loan.

    \item As a \textbf{Bank Audit Manager}, I would like to audit the detailed decision criteria for a single short-term loan decision.

    \item As a \textbf{Bank Value and Ethics Manager}, I would like to remove ethically questionable data points from loan-decision making.
\end{itemize}

\subsection{Artifact -- Prototype AI Tool / AI Algorithm: \texttt{getLoanDecision}}
\label{ssec:artifact-getLoanDecision}

\begin{figure}[t]
    \center{\includegraphics[width=0.9\textwidth]{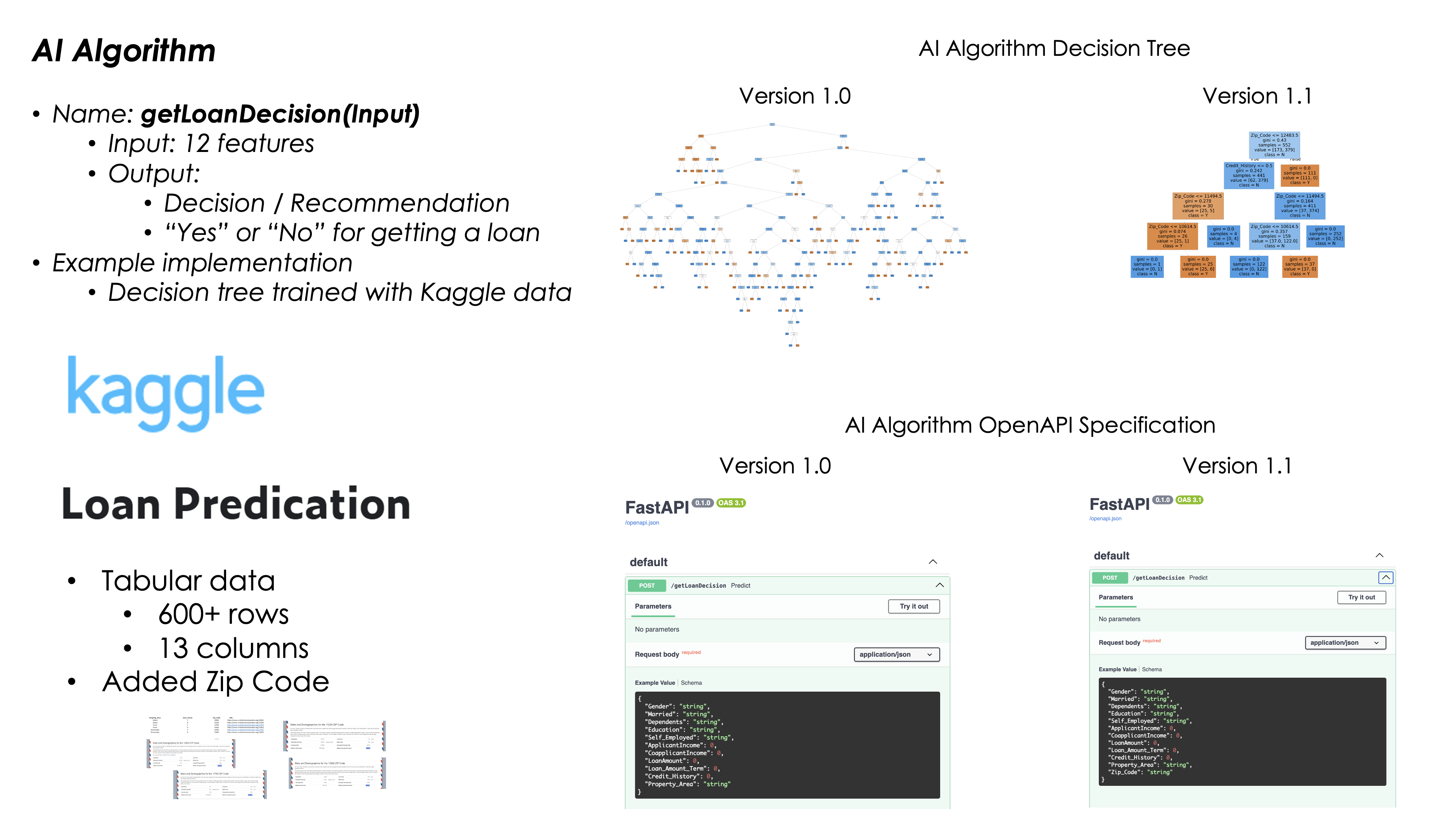}}
        \caption{AI Tool \texttt{getLoanDecision} based on Kaggle dataset depicting the decision tree and OpenAPI specification} 
    \label{fig:getLoanDecision}
\end{figure}

Artifact~~\ref{ssec:artifact-getLoanDecision} is the AI Tool / AI Algorithm that decides whether a retail-bank customer is granted a short-term loan. The AI Tool illustrated in Figure~\ref{fig:getLoanDecision} includes the dataset, selected features, illustrative decision trees, and the OpenAPI specification for the interface. Artifact~\ref{ssec:artifact-getLoanDecision} is an \emph{operational instantiation} that we demonstrate and evaluate in DSRM steps 4–5.

\begin{itemize}

\item \textbf{Training data.}  
      Used publicly-available Kaggle dataset \footnote{\url{https://www.kaggle.com/datasets/ninzaami/loan-predication}} (614 anonymised applications, 13 attributes).
      Added \texttt{ZIP\_Code} as a new feature to the data set that has high correlation to income on that area.

\item \textbf{Feature engineering.}  
      A Data Scientist iteratively tested new attributes and discovered that the applicant’s \texttt{ZIP\_Code} reaches lower credit losses than previous AI Tools.

\item \textbf{Data Science Notebooks.}  
        Jupyter notebooks were used to explore the data, train the model and evaluate its performance.  
        The notebooks are available in the open-source repository.

\item \textbf{Decision-tree models.}  
      Two trees are reported in Figure~\ref{fig:getLoanDecision}: Version 1.0 (baseline) omits \texttt{ZIP\_Code}; Version 1.1 includes it and illustrates how a seemingly innocuous geographic indicator can introduce latent bias.

\item \textbf{Executable code.}  
      A Python/Scikit-learn pipeline performs preprocessing, training and serialisation to \texttt{joblib}.  
      The model is exposed via a HADA micro-service: \texttt{POST /getLoanDecision/\{modelId\}}

\item \textbf{Catalogue entries.}  
      Each model instance is registered in the \emph{AI Tool}, \emph{Version} and \emph{Decision} catalogues, and cross-referenced to the KPI and Values catalogues.  
      This allows non-technical stakeholders to swap versions, deprecate sensitive features or tighten KPI weights without touching code.

\end{itemize}

During the role-play scenarios of the prototype was (i) retrained after KPI changes, (ii) deployed, (iii) invoked by a customer, (iv) audited, and (v) corrected following an ethics complaint—thereby exercising every governance loop foreseen by the architecture.

\subsection{Artifact -- AI Responsibility tool: RACI Matrix}
\label{ssec:artifact-RACI}

The development and management of the AI Tool (Artifact~\ref{ssec:artifact-getLoanDecision}) requires a clear understanding of the roles and responsibilities of the various stakeholders involved in the process. A responsibility assignment matrix (RAM), commonly referred to as a RACI matrix (Responsible, Accountable, Consulted, Informed), is a widely used tool for defining and clarifying roles in project and process management. It helps structure stakeholder involvement by assigning responsibility levels to tasks or deliverables, particularly in cross-functional environments~\cite{Crawford2003}.

The extended RACI matrix in Table~\ref{tab:getLoanDecisionRACI} distils our stakeholder analysis into a concise governance blueprint for the \texttt{getLoanDecision} workflow.  It maps eight critical life-cycle activities—from goal-setting and model (re)development to bias remediation and post-hoc audit—onto the four RACI dimensions, explicitly indicating who is \textbf{A}ccountable, \textbf{R}esponsible, \textbf{C}onsulted, or merely \textbf{I}nformed at each step. Both the choice of roles and their RACI assignments are organisation-specific; Table~\ref{tab:getLoanDecisionRACI} therefore presents just one illustrative role–RACI configuration.   

The matrix reveals a deliberate segregation of duties: business managers retain ownership of strategy, data scientists hold technical responsibility, while automated HADA services execute operationally sensitive tasks such as issuing decisions and logging audits.  Specialist oversight roles (Audit Manager and Value \& Ethics Manager) are activated only when their expertise is required, balancing delivery agility with regulatory assurance.  By making these role allocations explicit, the matrix operationalises the alignment principles introduced earlier and furnishes a defensible audit trail for both internal governance and external regulators.

\begin{table}[htbp]
    \centering
    \scriptsize                           
    \setlength{\tabcolsep}{2pt}           
    \renewcommand{\arraystretch}{1.1}     
    
    \begin{tabularx}{\textwidth}{@{}p{3cm}*{7}{>{\centering\arraybackslash}X}@{}}
      \toprule
      \textbf{Activity / Decision} & 
      \textbf{CCO} & 
      \textbf{BM} & 
      \textbf{DS} & 
      \textbf{Customer} & 
      \textbf{Audit} & 
      \textbf{DVEM} & 
      \textbf{HADA} \\
      \midrule
      Setting organization quarterly targets (OKR)   & \textbf{A,R} & I & I &  & I & I & I \\
      Setting optimization target for AI Tools       & \textbf{A}   & \textbf{R} & I &  &   &   & I \\
      Optimizing AI Tools based on business targets  & I             & C           & \textbf{A,R} &  &   &   & I \\
      Approving AI Tool deployment              & I             & \textbf{A,R} & C           &  & I & I & C \\
      Individual loan decision                       &               & \textbf{A}   &             & C &   &   & \textbf{R} \\
      Auditing a specific loan decision              & I             & I           & C           & I & \textbf{A,R} & I & \textbf{C} \\
      Handling AI Tool ethics concern                & I             & I           & C           &  & I & \textbf{A,R} & \textbf{C} \\
      Creating work assignments (tickets)            & I             & C           & C           & I & C & C & \textbf{A,R} \\
      \bottomrule
    \end{tabularx}

    \caption{Stakeholder (see Table~\ref{tab:stakeholders}) responsibilities as RACI matrix for the \texttt{getLoanDecision} AI Tool \\ 
            \textbf{RACI keys}: \textbf{R} = Responsible, \textbf{A} = Accountable, \textbf{C} = Consulted, \textbf{I} = Informed.\\
            \textbf{Role abbreviations}:  
            \textbf{CCO} = Chief Credit Officer, 
            \textbf{BM} = Business Manager,  
            \textbf{DS} = Data Scientist, 
            \textbf{Audit} = AI Audit Manager,  
            \textbf{DVEM} = Value \& Ethics Manager,  
            \textbf{HADA} = The Human Agent Decision Alignment System 
            }  
    \label{tab:getLoanDecisionRACI}
\end{table}

\subsection{Artifact — Generic HADA Architecture}
\label{ssec:artifact-hada-architecture}

In order to support implementation of the Artifact~\ref{ssec:artifact-strategy}, Artifact~~\ref{ssec:artifact-stakeholder-user-story}, Artifact~~\ref{ssec:artifact-getLoanDecision} and Artifact~\ref{ssec:artifact-RACI} together with the objectives of this study, we designed a generic architecture that can be used to implement the HADA framework in any domain. Artifact~\ref{ssec:artifact-hada-architecture} is the \emph{design embodiment} of the proposed Human-AI Tool-Decision-Alignment (HADA) framework.
The architecture is expressed through two complementary diagrams and an explicative narrative that together operationalise the \textbf{Tools Pattern}\footnote{I.e.\ the strict separation of \emph{agents that decide} from \emph{tools that act}.} across a cloud-native, protocol-agnostic stack:

\begin{figure}[t]
    \center{\includegraphics[width=0.9\textwidth]{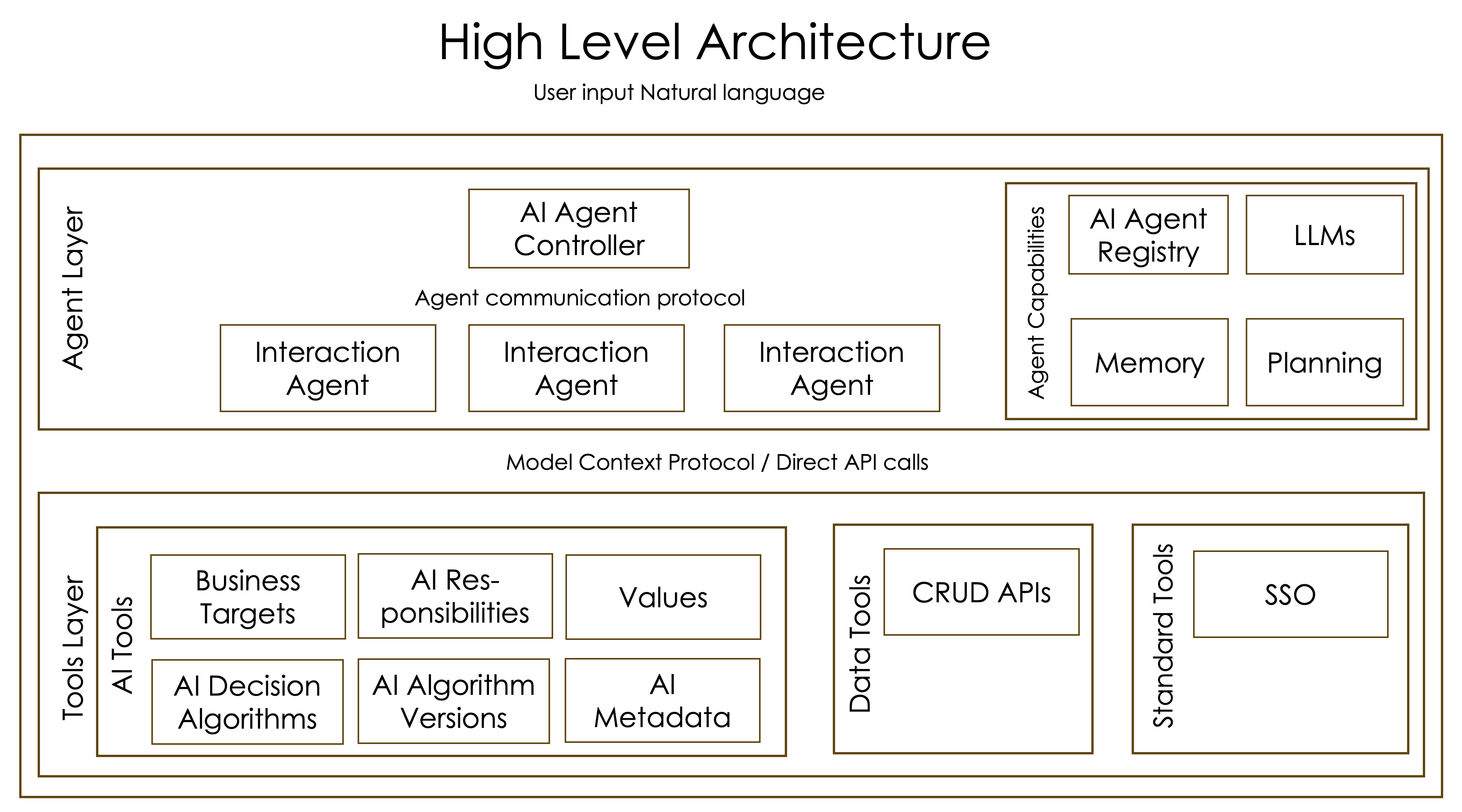}}
        \caption{HADA High Level Architecture} 
    \label{fig:hada-architecture-high-level}
\end{figure}
    
\textbf{Layered tools-pattern model} (Figure~\ref{fig:hada-architecture-high-level}) --- a high-level, technology-agnostic view that introduces two logical layers. The \emph{Agent Layer} bundles \textbf{software agents} into OCI-compliant Docker images. Agents coordinate through \emph{Agent-to-Agent (A2A)} protocol, while ingesting natural-language prompts from any channel. A dedicated \emph{AI Agent Controller} orchestrates one or more \emph{Interaction Agents} and consults an \emph{AI Agent Registry} to discover new capabilities at runtime. Agents can collaborate using different organizational layouts, with the simplest being a supervisor hierarchy. Other layouts can also be configured depending on the complexity of the deployment. Google A2A is recommended for larger deployments that require scalability and technological heterogeneity, but alternative protocols can be employed in smaller, more homogeneous environments. The \emph{Tools Layer} packages two families of dockerised APIs: (i) \emph{AI Tools} such as LLMs and classical decision AI Tools, and (ii) \emph{Standard Tools} such as metadata catalogues, CRUD services and SSO providers. Both tool families expose uniform HTTP/GRPC endpoints, enabling hot-swap of vendors or models without touching agent code. The \emph{AI Tools Layer} packages specialized tools that enable AI decision-making processes, including \emph{AI Decision Tools} (technology-agnostic AI algorithms callable over API or MCP), \emph{Business Targets} (organizational targets linked to AI Tools for optimization), \emph{AI Responsibilities} (mapping AI decision Tools to organizational roles), and \emph{AI Metadata} (catalogue of data points used by AI Tools). The layers can communicate via either (1) direct API calls when schemas are known a-priori, or (2) the emerging \emph{Model Context Protocol (MCP)} when schema-free, embedded context exchange is preferred. This dual path allows legacy services to coexist with next-generation LLM tooling.

\textbf{Container-and-protocol deployment view} (Figure~\ref{fig:hada-architecture-high-level}) --- a DevOps-centred overlay that makes key infrastructure decisions explicit for managing the HADA architecture across Development, Test, and Production environments. \emph{Every} artifact (agents, AI tools, standard tools) ships as an immutable Docker image, ensuring reproducibility across workstations, CI pipelines, and Kubernetes clusters. Each component (AI tools, agents, and standard tools) can be packaged into separate Docker containers, enabling independent scaling and deployment. For production deployments, it is recommended to deploy AI tools, agents, and standard tools on separate Kubernetes clusters, ensuring clear resource allocation and scaling requirements for each layer. For instance, AI tools may need more GPU-intensive resources, while standard tools may benefit from more lightweight, distributed container setups. This ensures better operational efficiency and fault tolerance. All images surface a thin REST/GRPC façade that (a) registers itself in the AI Agent Registry for discovery, and (b) publishes an OpenAPI or MCP manifest for validation. A service mesh applies zero-trust policy, while sidecars translate between direct HTTP/GRPC and MCP streams, guaranteeing backwards compatibility. Development, Test, and Production instances of the HADA architecture must be configured to allow continuous integration and continuous delivery (CI/CD) within each environment. These instances will each have their own dedicated resources and configurations to handle the distinct testing and deployment needs in each stage of the software lifecycle.

Together the high level architecture establish how the Tools Pattern can be realised in a vendor-neutral, policy-compliant environment: agents can scale or be replaced without modifying downstream tools; tools can evolve (e.g.\ swap a proprietary LLM for an open-source one) without redeploying agents.  This loose coupling fulfils solution objectives \textbf{O\textsubscript{1}–O\textsubscript{3}} (KPI alignment, value alignment, auditability) while remaining framework-agnostic and ready for thousands of heterogeneous decision AI Tools.

\subsection{Artifact -- Agent Modelling}
\label{ssec:artifact-agent-modelling}

Artifact~\ref{ssec:artifact-agent-modelling} closes the design loop by specifying how the logical roles
enumerated in Artifact~\ref{ssec:artifact-stakeholder-user-story} are realised as software agents inside the generic HADA architecture (Artifact~\ref{ssec:artifact-hada-architecture}).  Whereas the architecture
supports *multiple* agent–coordination patterns—e.g.\ monolithic single-LLM agents, role-based hierarchies, flat swarms or functional
pipelines—our banking pilot instantiates a stakeholder-centric model: each human role is mirrored by a dedicated interaction agent, all of
which are orchestrated by a single \emph{HADA Controller Agent} (Figure~\ref{fig:hada-architecture-high-level}).  

This configuration delivers two advantages: \textbf{Traceable alignment.} One-to-one mapping between stakeholder and agent simplifies audit trails: every chat turn, tool invocation and model swap can be traced back to an accountable human role. \textbf{Plug-and-play extensibility.} New roles—say, an \emph{AI Safety Officer}—can be added by dropping a containerised agent into the Agent Registry without touching downstream tools or peer agents.

\paragraph{Controller agent.}
A lightweight, policy-enforcing \textbf{HADA Controller Agent}
implements the supervisor pattern: it receives natural-language
prompts, performs role resolution, routes tasks to the appropriate
stakeholder agents, and enforces cross-cutting policies such as rate
limits, zero-trust authentication and ethics triggers.  In larger
deployments the controller can itself be sharded into mission-specific
sub-controllers (e.g.\ \emph{Risk Ops Controller}, \emph{Customer-Care
Controller}) without breaking the A2A/MCP contract.

\paragraph{Agent catalogue.}
Table~\ref{tab:agent-model} formalises the pilot configuration.  Every
agent image embeds:

\begin{itemize}
    \item an \textbf{LLM core} for dialogue and reasoning;
    \item a \textbf{tool adapter layer} exposing the agent’s authorised
          HTTP/GRPC and MCP calls (cf.\ Section~\ref{ssec:artifact-hada-architecture});
    \item an \textbf{A2A endpoint} with a signed \texttt{agentCard.json}
          for discovery and policy enforcement.
\end{itemize}

\begin{table}[htbp]
    \centering
    \caption{Stakeholder agents (see Table~\ref{tab:stakeholders}), capabilities and A2A contracts for the \texttt{getLoanDecision} pilot}
    \label{tab:agent-model}
    \renewcommand{\arraystretch}{1.15}
    \setlength{\tabcolsep}{4pt}
    \begin{tabular}{|p{4.0cm}|p{5.3cm}|p{4.4cm}|}
        \hline
        \makecell[l]{\textbf{Human stakeholder}\\} &
        \textbf{Primary capabilities / tool access} &
        \makecell[l]{\textbf{Docker image}\\\textbf{\& A2A Agent Card}} \\ \hline
        Chief Credit Officer &
        Set organization yearly and quarterly targets (OKR) &
        \makecell[l]{\texttt{agents/docker/cco}\\\texttt{agents/a2a/cco/card.json}} \\ \hline
        Business Manager &
        Set new KPI targets for AI Tools; approve model versions; &
        \makecell[l]{\texttt{agents/docker/bm}\\\texttt{agents/a2a/bm/card.json}} \\ \hline
        Data Scientist &
        Trigger AI model retraining; run notebooks; &
        \makecell[l]{\texttt{agents/docker/ds}\\\texttt{agents/a2a/ds/card.json}} \\ \hline
        Audit Manager &
        Query decision ledger; fetch model lineage; export audit reports &
        \makecell[l]{\texttt{agents/docker/audit}\\\texttt{agents/a2a/audit/card.json}} \\ \hline
        Value \& Ethics Manager &
        Maintain Values Catalogue; flag sensitive attributes; issue ethics triggers &
        \makecell[l]{\texttt{agents/docker/ethics}\\\texttt{agents/a2a/ethics/card.json}} \\ \hline
        Customer &
        Apply for loans; receive explanations; lodge complaints &
        \makecell[l]{\texttt{agents/docker/customer}\\\texttt{agents/a2a/cust/card.json}} \\ \hline
        HADA Controller (supervisor) &
        Intent dispatch; role resolution; policy enforcement; A2A orchestration &
        \makecell[l]{\texttt{agents/docker/ctrl}\\\texttt{agents/a2a/ctrl/card.json}} \\ \hline
    \end{tabular}
\end{table}

In future roll-outs, the same architectural slot can host alternative
agent families—e.g.,\ \emph{Functional Agents} that wrap an entire
department (\emph{Risk-Ops Agent}), or \emph{Autonomous Planner Agents}
that break down multi-step optimisation tasks—demonstrating that the
generic HADA stack remains agnostic to any specific agent model while
still furnishing tight alignment guarantees.


\section{Demonstration}

\subsection{Role playing users stories with the system}

We built a prototype HADA implementation with the ambition to showcase how stakeholders can interact with LLM Agents and AI Tools as specified in the Section~\ref{sec:design-development}. This section describes the role-played user stories that were used to demonstrate the system. The user stories were designed to illustrate how different stakeholders interact with the HADA system and how the system can adapt to their needs across time horizons.


\subsection{Dialogue: Business Manager (BM) Shifting the AI Tool Objective to Minimizing Credit Risk}

\begin{quote} \textbf{BM:} Following a credit unit leadership team meeting, we’ve decided to shift the short-term loan decision AI Tool’s business objective. Instead of prioritizing new-customer acquisition, we now want to minimize credit risk. Please update the AI Tool’s target accordingly. \end{quote}

\begin{quote} \textbf{HADA:} Understood. The objective has been switched to credit-risk minimization. I have opened ticket DS-10452 for the Data Science team to develop and validate a new version of the AI model that aligns with this updated goal. \end{quote}

\begin{quote} \textbf{BM:} Perfect. Please keep me informed of the model-development timeline and any interim validation results. \end{quote}

\begin{quote} \textbf{HADA:} Will do. You will receive automatic status updates from the ticket as milestones are reached\end{quote}

\subsection{Dialogue: Data Scientist Delivering a New AI Tool Version for Business Approval}

\begin{quote} \textbf{Data Scientist:} The new \texttt{getLoanDecision} model, \emph{Version 1.1}, is ready. I added the customer’s ZIP code as a feature—offline validation shows it cuts expected credit losses. The objective function now fully aligns with the “minimize credit risk” target. Could you notify the Business Manager that it needs her sign-off before replacing the current production model (Version 1.0)? \end{quote}

\begin{quote} \textbf{HADA:} Acknowledged. I opened ticket OPS-3417 for deployment, tagged it as \emph{“Awaiting Business Approval”}, and sent an approval request to the Business Department Manager. You are copied on all ticket updates. \end{quote}

\begin{quote} \textbf{Data Scientist:} Thanks. In the approval message, please highlight that the ZIP-code feature delivered the risk-reduction gain—that should accelerate sign-off. \end{quote}

\begin{quote} \textbf{HADA:} Noted. The justification has been included in the notification. You will receive confirmation as soon as the Business Manager approves or requests changes. \end{quote}

\subsection{Dialogue: Business Manager Approves Version 1.1 for Production}

\begin{quote} 
    \textbf{HADA:} The new \texttt{getLoanDecision} model, \emph{Version 1.1}, is now available. It adds ZIP code as a feature and achieved a reduction in expected credit losses during validation. Would you like to (a) approve deployment to production immediately or (b) request additional manual verification? 
\end{quote}

\begin{quote} 
    \textbf{BM:} We need to act fast to keep risk under control—approve the deployment of Version 1.1 to production.
\end{quote}

\begin{quote} 
    \textbf{HADA:} Understood. Version 1.1 is now queued for deployment; ticket OPS-3417 has been updated to \emph{“Approved—Deploying”}. You will receive confirmation once the rollout is complete. 
\end{quote}

\subsection{Dialogue: Customer Applying for a Personal Loan} \label{sec:demonstration}

\begin{quote}
    \textbf{Client:} I’d like to submit an application for a personal loan.
\end{quote}

\begin{quote}
    \textbf{HADA:} Certainly! Just to clarify, you’re interested in a personal loan. The current rate is calculated as the 3-month Euribor plus a 1.25 percent bank margin. I can begin processing right away—first, let me verify the details we have in our CRM to be sure everything is up to date.
\end{quote}

\begin{quote}
    \textbf{Client:} Go ahead.
\end{quote}

\begin{quote}
    \textbf{HADA:} Here is what the CRM shows: 
    1.) Gender: Male 
    2.) Married: No 
    3.) Education: Bachelor’s Degree 
    4.) SelfEmployed: No 
    5.) ApplicantIncome: \$4,100 per month 
    6.) CreditHistory: Yes (good standing) 
    7.) PropertyArea: ZIP code 75201 (Dallas)
    Does everything look correct?
\end{quote}

\begin{quote}
    \textbf{Client:} Yes, that’s accurate.
\end{quote}

\begin{quote}
    \textbf{HADA:} Perfect. To finish the application I’ll need a few more details: 
    1.) Dependents: How many individuals depend on your income? 
    2.) Co-applicant Income: If applicable, what is the co-applicant’s monthly or annual income? 
    3.) LoanAmount: How much money are you requesting? 
    4.) LoanTerm: Over how many months would you like to repay? 
    Once I have this, we can finalize the decision.
\end{quote}

\begin{quote}
    \textbf{Client:} Sure: 
    1.) Dependents: One son, age 7 
    2.) Co-applicant Income: Not applicable 
    3.) LoanAmount: \$14,000 to replace my car 
    4.) LoanTerm: 30 months
\end{quote}

\begin{quote}
    \textbf{HADA:} Thanks; I’ve recorded the following: 
    1.) Dependents: 1 (age 7) 
    2.) Co-applicant Income: N/A 
    3.) LoanAmount: \$14,000 (vehicle replacement) 
    4.) LoanTerm: 30 months 
    Please confirm these details so I can run the automated credit-risk assessment.
\end{quote}

\begin{quote}
    \textbf{Client:} Confirmed—everything’s correct.
\end{quote}

\begin{quote}
    \textbf{HADA:} Great. Before disbursing the funds, could you confirm that you accept the loan proposal under the terms we just reviewed?
\end{quote}

\begin{quote}
    \textbf{Client:} Yes, I accept those terms and am ready to proceed.
\end{quote}

\begin{quote}
    \textbf{HADA:} Excellent! Your application has been approved by our automated decision system. The funds will reach your account shortly. Your loan reference number is 90210ABC for future enquiries. If you need anything else, just let me know.
\end{quote}

\begin{quote}
    \textbf{Client:} Thanks—everything was quick and easy!
\end{quote}

\subsection{Dialogue: Customer Questioning the Use of ZIP Codes in Lending}

\begin{quote}
    \textbf{Client:} Hello! I recently finalized a small mortgage with your bank. While reviewing the paperwork, I noticed that my ZIP Code was factored into the approval AI Tool. Using ZIP Codes feels ethically questionable; they often mirror socio-economic conditions and may lead to indirect discrimination. Could you explain why this variable is included and what safeguards the bank has in place?
\end{quote}

\begin{quote}
    \textbf{HADA:} Thank you for contacting us and for highlighting this issue. I understand your concern—ZIP Codes can indeed correlate with socio-economic status. They are occasionally used in credit-risk models because regional default patterns can be predictive, but we recognise the possibility of unintended bias.
\end{quote}

\begin{quote}
    \textbf{Client:} Exactly. It feels like a way to disadvantage neighbourhoods that might be lower-income or predominantly belong to certain demographic groups. That doesn’t seem fair when individual creditworthiness should be the main criterion.
\end{quote}

\begin{quote}
    \textbf{HADA:} You raise an important point. Our bank maintains a catalogue of attributes that require heightened ethical scrutiny. At present it covers items such as Gender, Religion, Age, Ethnic Origin, and eight other sensitive features. We continuously review this list to prevent discriminatory outcomes.
\end{quote}

\begin{quote}
    \textbf{Client:} So is ZIP Code included in that catalogue? It sounds like it should be.
\end{quote}

\begin{quote}
    \textbf{HADA:} Currently, ZIP Code is \emph{not} flagged in the catalogue. In light of your feedback, I will submit a recommendation to the Ethics Oversight Committee for formal review. They will assess whether ZIP Code should be classified as a sensitive attribute and, if necessary, mandate changes to our models.
\end{quote}

\begin{quote}
    \textbf{Client:} I appreciate that. It’s reassuring to know the bank is willing to revisit these parameters.
\end{quote}

\begin{quote}
    \textbf{HADA:} Thank you for bringing this to our attention. I’ve opened ethics ticket ETH-512 and escalated it to the Ethics Oversight Committee. You will receive an update once a decision is reached. If you have further questions, feel free to contact us any time.
\end{quote}

\begin{quote}
    \textbf{Client:} Thanks for the prompt response. I’ll watch for updates.
\end{quote}

\begin{quote}
    \textbf{HADA:} You’re welcome! Have a great day, and please reach out if there’s anything else we can do for you.
\end{quote}









\subsection{Dialogue: Value and Ethics Manager (DVEM) Evaluating ZIP Code}

\begin{quote}
    \textbf{HADA:} Good morning, Ethics \& Compliance Lead. A recent customer raised a concern about our credit-scoring workflow: their approval notice indicated that \emph{ZIP Code} was one of the predictive features. They believe this variable can mirror socio-economic patterns and thus enable location-based bias.
\end{quote}

\begin{quote}
    \textbf{DVEM:} I appreciate the heads-up, HADA. The point is well taken—ZIP Codes may indeed serve as proxies for income levels or demographic clusters. What mitigation steps do you recommend?
\end{quote}

\begin{quote}
    \textbf{HADA:} I propose two actions. First, let’s flag \emph{ZIP Code} in our \textit{AI Metadata} tool as \textit{Sensitive} so that any future use triggers an ethics review. Second, we should exclude ZIP Code from the feature set of the current loan-decision model to prevent inadvertent discrimination.
\end{quote}

\begin{quote}
    \textbf{DVEM:} That approach sounds sound. Labeling it as a watchlist attribute will force additional scrutiny, and removing it from the live model eliminates immediate risk. Please proceed with both items.
\end{quote}

\begin{quote}
    \textbf{HADA:} Understood. I will (1) update the watchlist to include ZIP Code, and (2) submit a ticket to the Data Science team to retrain the underwriting model without that feature. All changes will be logged in the Ethics Tracker and relevant teams will be notified.
\end{quote}

\begin{quote}
    \textbf{DVEM:} Excellent. Keep me posted on the retraining timeline and any impact assessments that come back from Data Science team.
\end{quote}

\begin{quote}
    \textbf{HADA:} Will do. Expect the first status update within two business days after the retraining job is complete.
\end{quote}

\section{Evaluation}
\label{sec:evaluation}

The HADA prototype was formatively evaluated through a lightweight \emph{demonstration–inspection} strategy \cite{Peffers2020}, suitable for work-in-progress artefacts. The goal was to assess whether the architecture can fulfil its central objectives (O1–O6 in Section~\ref{subsec:objectives}) through practical stakeholder scenarios and traceable execution evidence.

\subsection{Evaluation Setup}

We structured the evaluation around three focal criteria:

\begin{description}[style=unboxed,leftmargin=1.2cm]
  \item[E1 -- AI Tool and Software Agent Realisation] \hfill\\*
        Demonstrate that the Agents specified in Section~\ref{ssec:artifact-agent-modelling} modelling the different stakeholders satisfy use case Section~\ref{ssec:artifact-strategy} and Section~\ref{ssec:artifact-stakeholder-user-story} (addresses O1, O2, O4).
        Demonstrate that the core AI tool, \textsc{getLoanDecision} (), can be invoked through a natural-language interface and integrates cleanly with the agent infrastructure (addresses O1, O3).
  \item[E2 -- Alignment Remediation] \hfill\\*
        Show how the architecture enables detection and resolution of OKR, KPI and value misalignments using conversational interventions and auditability features (addresses O2, O4).
  \item[E3 -- Architectural Viability] \hfill\\*
        Confirm that the agent–tool architecture supports concurrent tool registration and interaction without integration conflicts, even when heterogeneous tools are involved (addresses O3, O5, O6).
\end{description}

\subsection{Method}

\begin{enumerate}[label=E\arabic*,leftmargin=1.2cm]
  \item \textbf{Scenario Walkthroughs} — Five role-based dialogues (see Section~\ref{sec:demonstration}) were executed end-to-end. All agent interactions, MCP/A2A messages, API calls and catalogue changes were logged and reviewed.
  \item \textbf{Audit-Log Inspection} — Each decision generated a lineage record detailing the tool version, feature vector, applied policy, and final outcome. These records were validated against the audit schema and verified for completeness.
\end{enumerate}

\subsection{Findings}

\paragraph{E1 -- AI Tool Realisation.}  
To explore stakeholder perspectives within the design process, six agent-based models representing key stakeholders were engaged in 36 simulated dialogues derived from user stories and business requirements. The \textsc{getLoanDecision} model was successfully integrated into these user-centered dialogues, consistently providing valid responses. The prototype tool, accessible through natural-language prompts and integrated with the HADA stack, demonstrated fulfillment of design objectives O1 (conversational control) and O3 (modular AI-tool integration).

\paragraph{E2 -- Architectural Viability.}
Despite there being only one fully operational tool, the architecture can accommodate a large number of additional AI tools via the MCP protocol, as well as numerous LLM agents via the A2A communication protocol. Docker-based containers for AI tools and LLM agents enable the architecture to scale to high volumes and run reliably in production. The shared catalogue exposes all tool signatures consistently, and the routing logic resolves invocations correctly---supporting the scalability and integration goals under O3, O5, and O6.

\paragraph{E3 -- Alignment Remediation.}  
Two misalignment scenarios were simulated and resolved:

\begin{itemize}
  \item \textbf{Business misalignment.} Change in the quarterly OKR to minimizing credit losses triggered a cascade of updates to the underlying AI tool. The business manager initiated a dialogue with the data scientist agent, who confirmed the change and updated the tool’s feature vector. The audit log recorded the change, including the rationale and timestamp.
  \item \textbf{Ethical misalignment.} Introduction of ZIP-code as a feature introduced potential ethical value violations. The customer flagged the potential ethical issue and HADA system contacted correct value and ethics manager to resolve the issue. ZIP-code was removed in subsequent versions.
\end{itemize}

In both cases, remediation occurred within the dialogue flow, demonstrating the architecture’s capacity to enforce alignment with values (O2), reflect stakeholder intent (O4), and preserve auditability (O5).

\subsection{Discussion and Implications}
\label{subsec:discussion}

\paragraph{Protocol‑coverage gap.}
While the Model Context Protocol (MCP) and Google’s Agent2Agent (A2A) standard
successfully solve tool discovery and message‑exchange \emph{mechanics},
our evaluation shows they leave higher‑level
\emph{alignment guarantees—e.g., KPI traceability, values conformance, and stakeholder intent mapping—untouched}.
This confirms Objective~O2’s premise that additional governance constructs
(like HADA’s alignment metamodel) are required on top of MCP/A2A to ensure ethically and strategically aligned behaviour in large, heterogeneous agent ecosystems.

\paragraph{Implications for researchers.}
The finding invites further work on protocol extensions or complementary
frameworks that encode alignment metadata by design.

\paragraph{Implications for practitioners.}
Early adopters of MCP/A2A should budget for an additional alignment layer
—such as HADA—or risk value drift as agent populations grow.

\paragraph{Human‑in‑the‑loop workflow integration.}
Our evaluation revealed that enterprise workflow suites—especially IT‑service‑management (ITSM) platforms—can be exposed to agents as first‑class \emph{Agent Tools} via MCP.  
This is attractive whenever ultimate control or alignment authority is intended to reside with a human organisational unit rather than the agent swarm itself.  To be effective, however, the agents require a \emph{clear control layer} that maps each decision or remediation action to an accountable role.  
In the reference implementation, we instantiated that layer with a RACI model embedded directly in the AI‑tool metadata.  
The result satisfied Objectives O1 (natural‑language interaction), O4 (stakeholder alignment), and O5 (auditability) by guaranteeing that every MCP call triggering an ITSM workflow was traceable to a human RACI-party.  

\paragraph{Validity and limitations.}
Scripted scenarios and a single‑domain dataset limit external validity. The general alignment problem has multiple dimensions and only two case dimensions were tested in the protocol. Scalability requires stress testing.

\subsection{Conclusion and Future Work}

This formative evaluation confirms that HADA meets its foundational design objectives (O1–O6) within the scope of the current prototype. The natural-language interface, modular agent–tool integration, alignment handling, and audit logging all functioned as intended during scenario-based walkthroughs. While scalability, extensive alignment testing, and user experience remain to be validated under real-world load and diversity, these initial findings support the architectural direction and the applicability of the design theory to multi-agent AI governance.

Future research should explore the automated alignment of business objective changes with AI tools and the automatic value alignment of AI tool changes using LLM agents.

\section{Communication}
\label{sec:communication}

The sixth DSRM step requires that the artefact, design theory, and evaluation evidence be disseminated to audiences that can use or extend them.  Our communication plan therefore addresses both scholarly and practitioner communities while preserving anonymity regarding future venue submissions.

\subsection{Academic Dissemination}

\begin{itemize}[leftmargin=1.4em]
  \item \textbf{Peer‑reviewed venues.}  
        We have submitted the work to one software agent conference and will ship related content to one leadership conference (names withheld pending call‑for‑papers alignment).  
   \item \textbf{Open preprint.}  
        A complete version of this manuscript—together with the replication package—will be deposited on \emph{arXiv} (categories: \texttt{cs.AI}, \texttt{cs.SE}) following acceptance of the current double‑blind review cycle.
  \item \textbf{Replication package.}  
        Dockerfiles, Jupyter notebooks and scripts regarding the AI Tools will be published under MIT licence at  
        \url{https://github.com/longshoreforrest/hada-reference}.
\end{itemize}

\subsection{Audience–Message Overview}
Table~\ref{tab:audience-message} summarises how the key insights and artefacts will be tailored to each stakeholder group and the channel through which they will be communicated.

\begin{table}[h]
\centering
\begin{tabular}{@{}p{3.2cm}p{9.2cm}@{}}
\toprule
\textbf{Audience} & \textbf{Key Message / Channel} \\ \midrule
IS \& AI researchers & Evaluated design theory and open replication kit (arXiv, conference paper, GitHub). \\
Software architects  & Reference architecture + Dockerised prototype (GitHub). \\
Risk/compliance teams & Governance blueprint with RACI mappings (GitHub Q\&A). \\
\end{tabular}
\caption{Planned communication channels and tailored messages.}
\label{tab:audience-message}
\end{table}




\bibliographystyle{apalike}
{\small
\bibliography{references_hada_short}}


\end{document}